\documentclass[10pt,twocolumn,letterpaper]{article}

\usepackage[pagenumbers]{iccv} 
\usepackage{multirow}
\usepackage{tikz}
\usepackage{xcolor}

\definecolor{Color1}{RGB}{0, 51, 255}  
\definecolor{Color2}{RGB}{0, 255, 0} 
\definecolor{iccvblue}{rgb}{0.21,0.49,0.74}
\usepackage[pagebackref,breaklinks,colorlinks,allcolors=iccvblue]{hyperref}

\newcommand*{\affaddr}[1]{#1}

\title{OmniHuman-1: Rethinking the Scaling-Up of One-Stage  
\\ Conditioned Human Animation Models}

\author{
Gaojie Lin$^{*}$ \quad 
Jianwen Jiang$^{*\dagger}$ \quad 
Jiaqi Yang$^{*}$ \quad 
Zerong Zheng$^{*}$ \quad 
Chao Liang \vspace{2mm} \\
\affaddr{ByteDance} \\
\small{\href{https://omnihuman-lab.github.io/}{\ttfamily https://omnihuman-lab.github.io/}}
}

\begin{document}

\twocolumn[{
\maketitle
\begin{center}
    \vspace{-10pt}
    \captionsetup{type=figure}
    \includegraphics[width=1\textwidth]{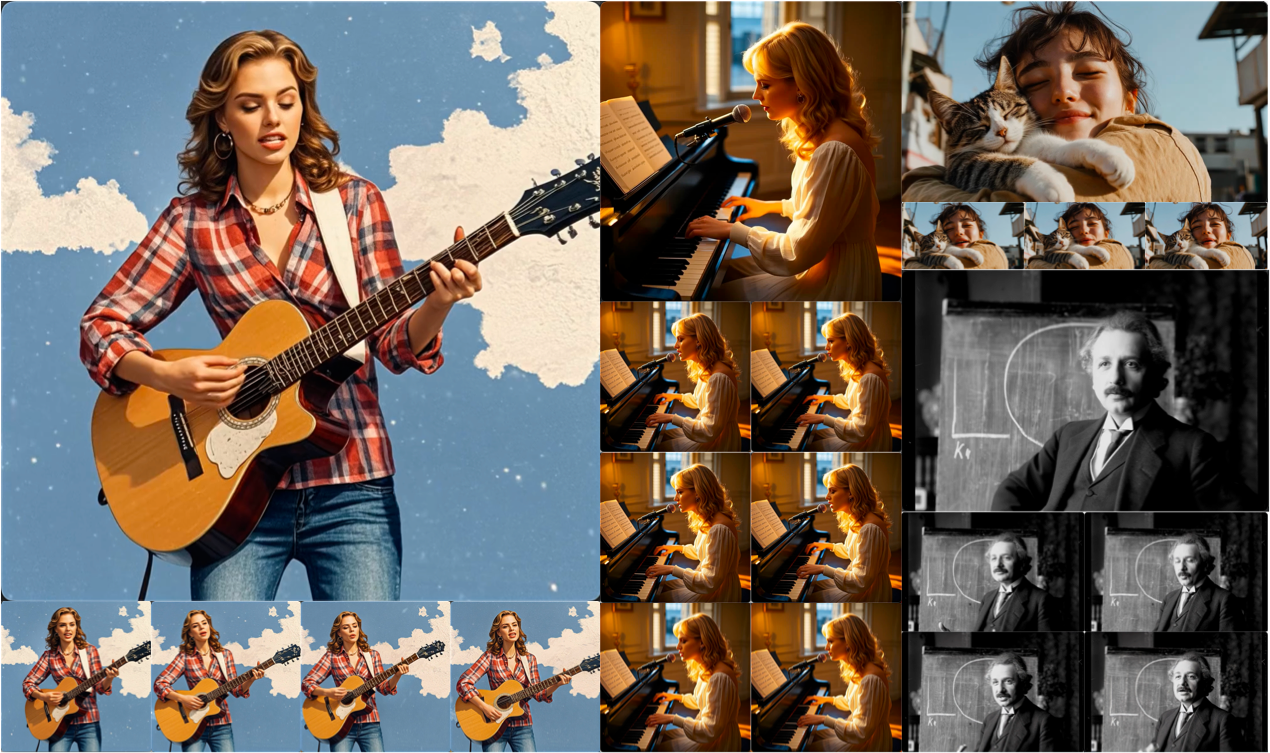}
    \captionof{figure}{\small \textbf{The video frames generated by OmniHuman based on input audio and image.} The generated results feature head and gesture movements, as well as facial expressions, that match the audio. OmniHuman generates realistic videos with any aspect ratio and body proportion, and significantly improves gesture generation and object interaction over existing methods, due to the data scaling up enabled by omni-conditions training.}
    \label{fig:vis1}
\end{center}
}\bigbreak
]

\footnotetext{$^*$Equal contributions}
\footnotetext{$^\dagger$Project lead and corresponding author: jianwen.alan@gmail.com}

\begin{abstract}
End-to-end human animation, such as audio-driven talking human generation, has undergone notable advancements in the recent few years. However, existing methods still struggle to scale up as large general video generation models, limiting their potential in real applications. In this paper, we propose OmniHuman, a Diffusion Transformer-based framework that scales up data by mixing motion-related conditions into the training phase. To this end, we introduce two training principles for these mixed conditions, along with the corresponding model architecture and inference strategy. These designs enable OmniHuman to fully leverage data-driven motion generation, ultimately achieving highly realistic human video generation. More importantly, OmniHuman supports various portrait contents (face close-up, portrait, half-body, full-body), supports both talking and singing, handles human-object interactions and challenging body poses, and accommodates different image styles. Compared to existing end-to-end audio-driven methods, OmniHuman not only produces more realistic videos, but also offers greater flexibility in inputs. It also supports multiple driving modalities (audio-driven, video-driven and combined driving signals). 
\end{abstract}    
\section{Introduction}
Since the emergence of the Diffusion Transformer-based (DiT) video diffusion models, the field of general video generation, including Text-to-Video and Image-to-Video~\cite{bar2024lumiere,svd,ayl,guo2023animatediff,zhou2022magicvideo,walt,wang2023modelscope,vdm,videogan,cvideogan,singer2022make,text2video,villegas2022phenaki,lin2025apt} has made significant progress in producing highly realistic video content. A key factor driving this advancement is the large-scale training data, typically formatted as video-text pairs. Expanding the training dataset enables DiT networks to learn motion priors for various objects and scenes, resulting in strong generalization capabilities during inference.

Building upon these pretrained video diffusion networks, end-to-end human animation models, especially for audio-driven talking human generation, have developed rapidly since last year~\cite{he2023gaia,tian2024emo,xu2024hallo,wang2024vexpress,chen2024echomimic,xu2024vasa,stypulkowski2024diffused,jiang2024loopy,lin2025cyberhost}. 
Despite achieving realistic results, these models are trained on highly filtered datasets to simplify the learning process, restricting their applicability to limited scenarios. 
For instance, most existing end-to-end audio-driven models are limited to facial or portrait images captured from a front-facing perspective with a static background. To date, no prior work has attempted to scale up training data for more generalizable human animation.

Scaling up human animation data may seem straightforward, but unfortunately it is not. Directly adding more data is not always beneficial for network training. 
Take audio-conditioned models as an example: audio is primarily associated with facial expressions and has little correlation with body poses, background motion, camera movement, or lighting changes. As a result, raw training data must be filtered and cropped to minimize the influence of these unrelated factors. Additionally, audio-conditioned models often undergo further data cleaning based on lip-sync accuracy, which is also important to stabilize training. 
Unfortunately, these processes discard a substantial amount of data, making dataset scaling a futile effort or difficult to achieve, despite much of the discarded data containing valuable motion patterns essential for training data expansion. For example, recent state-of-the-art methods mention that after rigorous cleaning, less than 10\% of the data is retained, rendering direct data scaling highly cost-ineffective.

In this paper, we address the challenges of scaling up audio-driven human animation data and models. Our key insight is that incorporating multiple conditioning signals beyond audios, such as text and pose, during training can significantly reduce data wastage. This approach offers two main advantages. On one hand, data that would otherwise be discarded for single-condition models (e.g., audio-driven) can be leveraged in tasks with weaker or more general conditions, such as text. Training on such data allows the model to learn more diverse motion patterns, mitigating the limitations imposed by data filtering. On the other hand, different conditioning signals can complement each other. For example, while audio alone cannot precisely control body poses, stronger conditions such as body poses can provide additional guidance. By integrating stronger conditioning signals alongside audio data, we aim to reduce overfitting and improve the generalization of generated results.

Building on the above considerations, we designed an omni-conditions training strategy, which follows two key training principles: (1) tasks with stronger conditioning can leverage those with weaker conditioning, along with their associated data, to scale up training, and (2) the stronger the conditioning signal, the lower its training ratio should be. To implement these strategies, we develop OmniHuman, a mixed conditioned human video generation model based on the advanced video generation architecture, DiT~\cite{dit,sd3}. Although OmniHuman mainly focuses on audio-driven human video generation, it is trained with three motion-related conditions, including text, audio, and pose, ranging from weak to strong. This approach effectively addresses the challenge of data scaling in end-to-end audio-driven frameworks, allowing the model to learn natural motion patterns from large-scale data and support various input forms.

In summary, we introduce OmniHuman, an audio-driven human video generation model that leverages our omni-conditions training strategy to integrate various motion-related conditions and their corresponding data. Unlike existing methods that reduce data due to strict filtering, our approach benefits from large-scale mixed-conditioned data, enabling OmniHuman to generate highly realistic and expressive human motion videos given a reference image and the driving audio. It adapts to various portrait types and aspect ratios, and significantly improves gesture generation, which is a longstanding challenge for previous methods. Additionally, it accommodates diverse image styles and background contents. To the best of our knowledge, OmniHuman is the first solution capable of audio-driven human video generation on input images with any body proportions and image styles, and also supports auxiliary pose driving.

\section{Related Works}
\subsection{Video Generation}
  In recent years, the advent of technologies such as diffusion models \cite{jonathan2020ddpm, song2021ddim, karras2022edm, song2020score, liu2022reflow} has propelled the capabilities of generative models to a practically usable level. The latest advancements in image generation \cite{sd3, chen2024pixartdelta} produce results that are almost indistinguishable from reality. Consequently, a growing number of studies \cite{zhou2022magicvideo, zeng2024pxldance, hong2022cogvideo, yang2024cogvideox, openai2024sora, kong2024hunyuanvideo, polyak2024moviegen} are shifting their focus toward the field of video generation.
  Early text-to-video works primarily centered on training-free adaptations of pre-trained text-to-image models \cite{singer2022make, wu2023tune, qi2023fatezero} or integrated temporal layers with fine-tuning on limited video datasets \cite{guo2023animatediff, zhou2022magicvideo, wang2023modelscope}. However, due to the lack of extensive data, the video generation quality of these methods often remains unsatisfactory. To better exploit scaling laws and push the boundaries of video generation models, recent works \cite{openai2024sora, yang2024cogvideox, kong2024hunyuanvideo, polyak2024moviegen} have optimized in three major areas. First, they have collected larger-scale, high-quality video datasets, with the data volume increasing to (O(100M)) clips of high-resolution videos. Second, they employ 3D Causal VAE \cite{yu20233DVAE} to compress both spatial and temporal features of video data, thereby enhancing video modeling efficiency. Third, the foundational model structure has transitioned from UNet to Transformer, improving the model’s scalability. Additionally, these works utilize meticulously designed progressive training recipes and datasets to maximize the model's potential. For example, \cite{polyak2024moviegen, kong2024hunyuanvideo} first pre-train on a large volume of low-resolution images and videos, leveraging data diversity to enhance the model's generalization capabilities. They then perform fine-tuning on a subset of high-resolution, high-quality data to improve the visual quality of generated videos. Large-scale data has significantly improved the effectiveness of general video generation. However, progress in the field of human animation synthesis remains relatively slow.
  
  \subsection{Human Animation}
  As an important task of video generation, Human Animation synthesizes human videos using human images and driving conditions such as audios or videos. Early  GAN-based methods \cite{siarohin2019fomm, zhao2022tps, siarohin2021mraa, jiang2024mobileportrait, wang2021facev2v} typically employ small datasets \cite{nagrani2017voxceleb, siarohin2019fomm, xie2022vfhq, zhu2022celebv} consisting of tens of thousands of videos to achieve video-driven in a self-supervised manner. With the advancement of Diffusion models, several related works \cite{Disco, aa, champ, shao2024human4dit, zhang2024mimicmotion} have surpassed GAN-based methods in performance while using datasets of similar scale. Instead of using pixel-level videos, these methods employ 2D skeleton, 3D depth, or 3D mesh sequences as driving conditions.
  Audio-driven methods used to focus on portrait \cite{adnerf,GeneFace,zhang2023sadtalker, tian2024emo, jiang2024loopy, hallo3, fada}. Despite some efforts \cite{VLogger, lin2025cyberhost, EchomimicV2, EMO2, diffted} to extend the frame to the full body, there are still challanges especially in hand quality. To bypass it, most approaches \cite{VLogger,  EchomimicV2, EMO2, diffted} adopt a two-stage hybrid driving strategy, utilizing gesture sequences as a strong condition to assist hand generation. CyberHost \cite{lin2025cyberhost} attempts to achieve one-stage audio-driven talking body generation through codebook design.
  Most notably, existing Human Animation methods typically focus on limited-scale datasets and limited-complexity structure, generally less than a thousand hours and 2B. Although FADA \cite{fada} employs a semi-supervised data strategy to utilize 1.4K hours of portrait videos, VLogger \cite{VLogger} meticulously collects 2.2K hours of half-body videos, and Hallo3 \cite{hallo3} initializes its weights derived from CogVideoX5B-I2V \cite{cogvideox}, their performance  does not exhibit the scaling law trends observed in other tasks such as LLMs \cite{ouyang2022training, touvron2023llama}, VLMs \cite{liu2024improved,bai2023qwen}, and T2I/T2V \cite{esser2024scaling, flux2023, kondratyuk2023videopoet}. Scaling effects in Human Animation  haven't been investigated effectively yet.

\begin{figure*}[t]
    \centering
    \includegraphics[width=\textwidth]{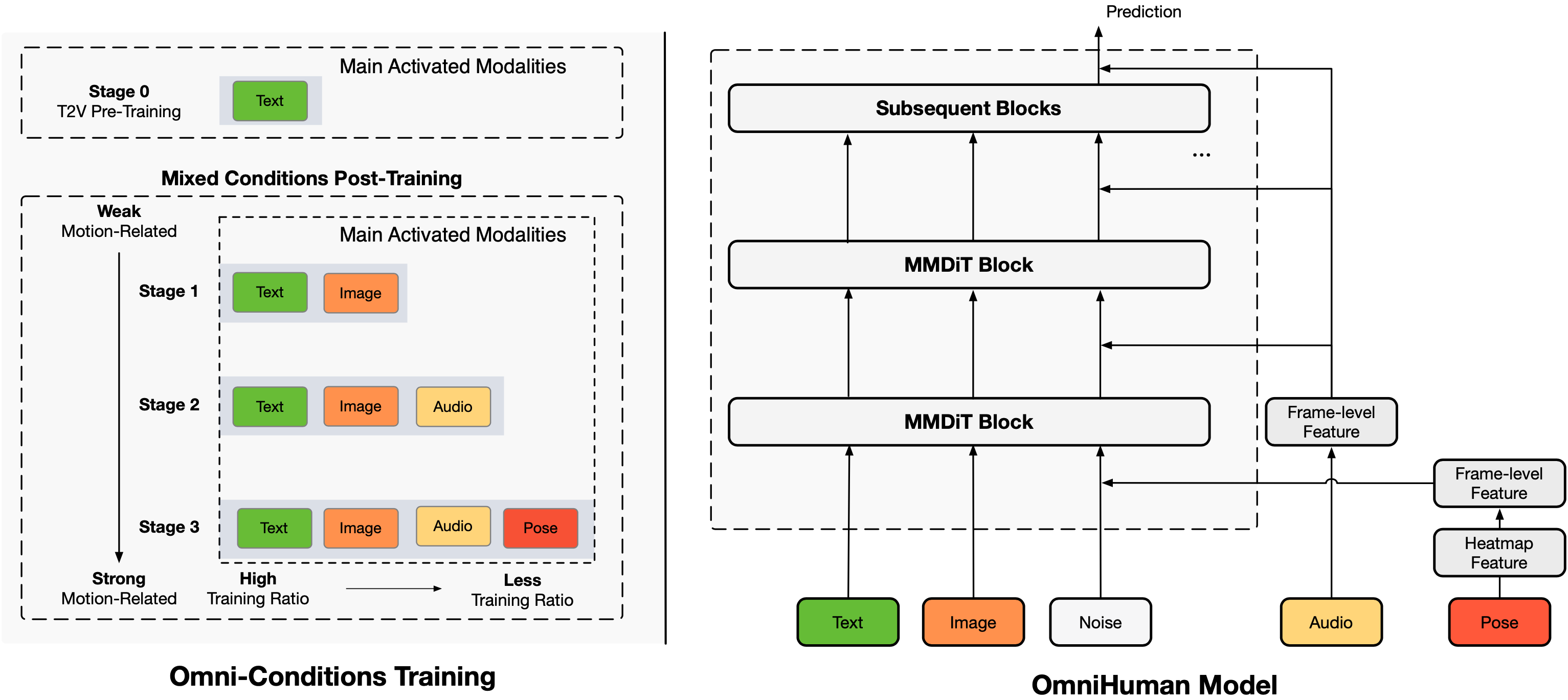}
    \caption{\small \textbf{The framework of OmniHuman.} It consists of two parts: (1) the OmniHuman model, which is based on the DiT architecture and supports simultaneous conditioning with multiple modalities including text, image, audio, and pose. To support long video continuation, we concatenate the latents of the last generated frames with noise latents, which are omitted for simplicity. (2) the omni-conditions training strategy, which employs progressive, multi-stage training based on the motion-related extent of the conditions. The mixed condition training allows the OmniHuman model to benefit from the scaling up of mixed data.}
    \label{fig:framework}
\end{figure*}
\section{Method}
\label{headings}

\subsection{Overview}

The goal of OmniHuman is to generate realistic human animation videos from a reference image and using one or more driving conditions, including text, audio and pose. Here we mainly focuses on audio-conditioned generation, and leverage the other two as auxiliary signals.
To achieve this,  our approach consists of two primary parts: the OmniHuman model, a multi-condition diffusion model, and the Omni-conditions Training Strategy. Figure~\ref{fig:framework} illustrates our approach. Our OmniHuman model is built upon an MMDiT~\cite{sd3,dit,lin2025apt,navit}-based text-to-video model, which is initially trained on general text-video pairs. We design different methods to integrate different driving signals, including audio features and skeleton map features, into the OmniHuman model (Sec.~\ref{sec:method:condition}). Then we employ a three-stage mixed condition training approach to progressively transform the diffusion model from a general text-to-video model to a multi-condition human video generation model (Sec.~\ref{sec:method:training}). As illustrated on the left of Figure~\ref{fig:framework}, these stages sequentially introduce the driving modalities of text, audio, and pose based on their motion correlation strength, from weak to strong, and balance their training ratios. This ultimately enables the model to be compatible with different modalities and benefit from large-scale data with mixed conditions. 

Note that OmniHuman operates in a compact latent space by using a causal 3D VAE~\cite{opensora} to project videos at their native resolution into compact latents. However, for simplicity, we omit this process in the following sections.

\subsection{Omni-Conditions Module Design}
\label{sec:method:condition}
For designing the condition injection module for OmniHuman, we adhere to a minimalist design principle, aiming to minimize modality-specific parameters. This ensures that multimodal interactions can be more extensively modeled within the shared MMDiT backbone, which contains the majority of model parameters. We categorize the roles of modalities in the human animation model into two types of conditions, as discussed below.

\textbf{Driving Conditions.} In OmniHuman, driving conditions include audio, pose, and text, with audio being the primary focus of this paper. For the audio condition, the wav2vec  model \cite{schneider2019wav2vec,baevski2020wav2vec2} is employed to extract multi-scale acoustic features \cite{jiang2024loopy, tian2024emo}. These features are subsequently compressed using an MLP to align with the hidden size of MMDiT and the framerate of videos (25 fps in our setting). The features of each frame are concatenated with audio features from adjacent timestamps to generate audio tokens for the current frame. As depicted in Figure~\ref{fig:framework}, these audio tokens are injected into each block of MMDiT through frame-wise cross-attention, enabling interaction between the audio tokens and the noisy latent representations. 

To incorporate the pose condition, we use a pose guider~\cite{aa} to encode the driving skeleton map sequence. The resulting pixel-aligned pose features are concatenated with those of adjacent frames to acquire frame-wise pose tokens. These pose tokens are then stacked with the noisy latent representations along the channel dimension and fed into the model for visual alignment and dynamic modeling. 

The text condition is processed with the original MMDiT's text branch, following the common practice of text-to-image and text-to-video frameworks~\cite{sd3,dit}.

\textbf{Appearance Conditions.} The appearance conditioning is incorporated to preserve the identity of the foreground character while ensuring natural background motion. Previous research often used a reference network approach \cite{jiang2024loopy,lin2025cyberhost,tian2024emo}, which involves a parallel, trainable copy of the entire diffusion backbone that integrates with the self-attention layers of the original denoising network. While effective, this method requires duplicating a full set of trainable parameters, posing scalability challenges as model size increases. 
In contrast, our OmniHuman adopts a simpler yet effective strategy for reference conditioning by reusing the original denoising DiT backbone to encode the reference image. Specifically, both the reference and noisy video latents are flattened into token sequences, packed together, and simultaneously fed into the DiT. This allows the reference and video tokens to interact via self-attention across the entire network. To distinguish between reference and video tokens, we modify the 3D Rotational Position Embeddings (RoPE) \cite{rope} in the DiT by zeroing the temporal component for reference tokens while leaving the RoPE for video tokens unchanged. Such an approach effectively incorporates appearance conditioning without adding extra parameters, maximizing the modeling of interactions between different modalities within the shared parameters. 

In addition to the reference image, to support long video generation, we use motion frames \cite{stypulkowski2024diffused} by simply concatenating their features with the noise tokens as in \cite{opensora}.

\subsection{Scaling up with Omni-Condition Training}
\label{sec:method:training}

Thanks to its multi-condition design, OmniHuman enables training across multiple tasks, including: (1) image and text to video, (2) image, text, and audio to video, and (3) image, text, audio, and pose to video. During training, different modalities are activated based on the available data, allowing a wider range of data to contribute to the training process and enhancing the model’s generative capabilities. Starting with a pretrained general text-to-video model, we follow two key training principles to scale up conditioned human video generation, progressively transforming the pretrained model into a mixed-conditioned human generation framework.

\textbf{Principle 1}: Stronger conditioned tasks can leverage weaker conditioned tasks and their corresponding data to scale up training data during model training. For instance, data excluded from audio-conditioned tasks due to filtering criteria, such as lip-sync accuracy and stability, can still be utilized in text- and image-conditioned tasks, as they meet the standards for weaker conditions.
Therefore, in Stage 1, we exclude the audio and pose conditions and train the model on the text+image-to-video task, maximizing the use of available training data.

\textbf{Principle 2}: The stronger the condition, the lower its training ratio should be. During training, stronger motion-related conditions, such as pose, generally provide more precise motion control compared to weaker conditions like audio, which may introduce ambiguity. When both conditions are present, the model tends to rely on the stronger condition, limiting the learning effectiveness of the weaker one.
To achieve a unified multi-condition human animation model, we balance this effect by assigning a higher training ratio to weaker conditions and adjusting their training order, thus ensuring they contribute meaningfully to motion generation.
To be more specific, we drop only the pose condition in Stage 2 and incorporate all conditions in the final Stage 3. The training ratios for text, audio, and pose are progressively halved when present, assigning higher gradient weights to more challenging tasks and preventing overfitting to a single condition during mixed-condition training.

In summary, Principle 1 allows us to significantly expand the training data, while Principle 2 ensures that the model fully leverages each motion-related condition’s strengths during mixed-condition training, enabling it to learn distinct motion generation capabilities and avoiding either overfitting or underfitting. By integrating both principles, OmniHuman effectively trains on mixed-conditioned data, benefiting from data scaling while achieving high-quality results.

\subsection{Inference Strategies}
For audio-driven scenarios, all conditions except pose are activated. For mixed driving with pose and audio, all conditions are activated. For pose-only driving, audio is disabled. In general, when a condition is activated, all conditions with a lower motion-related influence are also activated unless deemed unnecessary. Additionally, for all animation tasks, we use an image captioning model to obtain the text input.

During inference, to balance expressiveness and computational efficiency, we apply classifier-free guidance (CFG) specifically to audio and text, but not to the pose condition. The CFG scale is set to 6.5. We found that only conditions with relatively weaker motion-related influence, such as audio and text, require CFG. To maintain temporal coherence and identity consistency in long videos, the last five frames of the previous segment are used as motion frames for the following segment.

\begin{table*}[ht]
    \small
    \footnotesize
    \setlength{\tabcolsep}{2.5pt}
    \centering
    \caption{\small \textbf{Quantitative analysis of Omni-Conditions Training}. The upper and lower parts correspond to Principles 1 and 2 respectively.}
  \label{table:ablation}
  \centering
  \begin{tabular}{cccccccccccccccccc}
    \toprule
     \multirow{2}{*}{Methods} & \multicolumn{5}{c}{CelebV-HQ} & \multicolumn{5}{c}{RAVDESS}  & \multicolumn{7}{c}{CyberHost} \\
     \cmidrule(l){2-6}
     \cmidrule(l){7-11}
      \cmidrule(l){12-18}
     &IQA $\uparrow$ & ASE$\uparrow$ & Sync-C$\uparrow$ & FID$\downarrow$ & FVD$\downarrow$
     &IQA $\uparrow$ & ASE$\uparrow$ & Sync-C$\uparrow$ & FID$\downarrow$ & FVD$\downarrow$
     &IQA$\uparrow$  & ASE$\uparrow$ & Sync-C$\uparrow$  & FID$\downarrow$ & FVD$\downarrow$ & HKV & HKC$\uparrow$ \\
    \midrule
    Ref Img    &3.195 &1.963 &- &- &- &4.421 &2.578 &- &- &- &3.963 &2.634 &- &- &- &- &- \\
    $0\%$ T-Data      &3.855 &2.693 &4.299 &39.80 &47.86 &4.671 &3.053 &4.663 &18.20 &18.83  &4.262 &3.171 &6.465 &38.38 &45.77 &35.82 &0.871 \\
    $25\%$ T-Data   &3.758 &2.571 &3.311 &37.95 &47.04 &4.652 &3.011 &3.552 &17.03 &16.00 &4.236 &3.152 &5.039 &38.36 &43.69 &40.39 &0.877 \\
    $50\%$ T-Data  &3.717 &2.525 &3.696 &36.26 &46.22 &4.541 &2.913 &3.826 &16.92 &15.51  &4.217 &3.132 &5.591 &37.06 &37.95 &40.69 &0.872 \\
    $100\%$ T-Data &3.644 &2.452 &4.987 &36.01 &43.74 &4.445 &2.763 &5.092 &16.70 &15.13 &4.076 &2.959 &6.938 &33.27 &31.15 &43.54 &0.882 \\
    \midrule
    IA                       &3.644 &2.452 &4.987 &36.01 &43.74 &4.445 &2.763 &5.092 &16.70 &15.13 &4.076 &2.959 &6.938 &33.27 &31.15 &43.54 &0.882 \\
    IPA                      &3.458 &2.279 &2.788 &38.98 &44.70 &4.275 &2.681 &2.923 &21.20 &25.05 &3.945 &2.815 &4.879 &39.60 &36.03 &45.44 &0.822 \\
    IAP, A\textless P      &3.556 &2.370 &4.201 &38.73 &44.63 &4.298 &2.689 &4.752 &20.79 &17.18 &3.979 &2.869 &6.765 &34.26 &32.79 &40.99 &0.869 \\
    IAP, A\textgreater P   &3.602 &2.415 &4.934 &36.66 &43.36 &4.425 &2.796 &4.882 &17.43 &15.66 &4.055 &2.959 &6.951 &33.27 &31.08 &39.39 &0.886 \\
  \bottomrule
  \end{tabular}
\end{table*}

\section{Experiments}
\subsection{Implementation Details}
  \textbf{Dataset.} We use 18.7K hours of in-house human-related data for training, which are filtered based on aesthetics, image quality, motion amplitude, etc. (common criteria for video generation~\cite{chen2024panda}). Of this dataset, 13\% was selected using lipsync and pose visibility criteria, enabling audio and pose modalities. During training, the data composition was adjusted to fit the omni-condition training strategy. For testing, we conduct the evaluation following the portrait animation method Loopy \cite{jiang2024loopy} and the half-body animation method CyberHost \cite{lin2025cyberhost}. We randomly sampled 100 videos from public portrait datasets, including CelebV-HQ \cite{zhu2022celebv} (a diverse dataset with mixed scenes) and RAVDESS \cite {ravdess_dataset} (an indoor dataset including speech and song) as the testset for portrait animation. For half-body animation, we used CyberHost's test set, which includes a total of 269 body videos with 119 identities, encompassing different races, ages, genders, and initial poses.

  \textbf{Baselines.} To comprehensively evaluate OmniHuman's performance in different scenarios, we compare against portrait animation baselines including Sadtalker \cite{zhang2023sadtalker}, Hallo \cite{xu2024hallo}, Vexpress \cite{wang2024vexpress}, EchoMimic \cite{chen2024echomimic}, Loopy \cite{jiang2024loopy}, Hallo-3 \cite{hallo3}, and body animation baselines including DiffTED \cite{diffted}, DiffGest \cite{zhu2023taming} + Mimiction \cite{zhang2024mimicmotion}, CyberHost \cite{lin2025cyberhost}. 

  \textbf{Metrics.} For visual quality, FID \cite{heusel2017gans} and FVD \cite{unterthiner2019fvd} are used to evaluate the distance between the generated and labeled images and videos. We also leverage q-align \cite{wu2023q}, a VLM to evaluate the no-reference IQA(image quality) and ASE(aesthetics). For lip synchronism, we employ the widely-used Sync-C \cite{syncnet} to calculate the confidence between visual and audio content. Besides, HKC (hand keypoint confidence) \cite{lin2025cyberhost} and HKV (hand keypoint variance) \cite{lin2025cyberhost} are employed, to represent hand quality and motion richness respectively. 
  \par

   During training, we use a learning rate of $5 \times 10^{-5}$ with the AdamW optimizer, a gradient clip of 1.0, and a batch size of 256. The weight decay is set to 0.01. The training process is carried out on 400 A100 GPUs, with each phase lasting approximately 10 days.
\par
In the following sections, we first validate the effectiveness of the OmniHuman design through ablation studies, primarily using audio driving for comparison because it not only reflects mimicking ability but also better showcases motion generation capability. For efficiency, experiments are conducted at 480P resolution. Next, we present several visual results and applications to illustrate the generalization capability of OmniHuman. Finally, we compare our results at 720P resolution with current state-of-the-art methods to demonstrate OmniHuman's versatility across different tasks. It is important to note that our aim is to demonstrate that the OmniHuman paradigm better facilitates data scaling for more general human animation rather than focusing on performance in limited scenarios. This could potentially be an important training paradigm for human animation methods.

\begin{figure*}
    \small
\captionsetup{type=table}
    \centering
        {
            \caption{\textbf{Quantitative comparisons with audio-conditioned portrait animation baselines.}}
  \label{Tab1}
  \centering
  \begin{tabular}{ccccccccccc}
    \toprule
     \multirow{2}{*}{Methods} & \multicolumn{5}{c}{CelebV-HQ} & \multicolumn{5}{c}{RAVDESS} \\ 
     \cmidrule(l){2-6}
     \cmidrule(l){7-11}
     &IQA $\uparrow$ & ASE$\uparrow$ & Sync-C$\uparrow$ & FID$\downarrow$ & FVD$\downarrow$ &IQA $\uparrow$ & ASE$\uparrow$ & Sync-C$\uparrow$ & FID$\downarrow$ & FVD$\downarrow$  \\ 
    \midrule
         SadTalker \cite{zhang2023sadtalker} & 2.953 & 1.812 & 3.843 & 36.648 & 171.848 & 3.840 & 2.277 & 4.304 & 32.343 & 22.516  \\ 
         Hallo \cite{xu2024hallo} & 3.505 & 2.262 & 4.130 & 35.961 & 53.992 & 4.393 & 2.688 & 4.062 &19.826 & 38.471\\ 
         VExpress \cite{wang2024v} & 2.946 & 1.901 &3.547 & 65.098 & 117.868 & 3.690 & 2.331 & 5.001 & 26.736 &  62.388 \\ 
         EchoMimic \cite{chen2024echomimic} & 3.307 & 2.128 & 3.136 & 35.373 & 54.715 & 4.504 & 2.742 & 3.292 & 21.058 & 54.115 \\ 
    Loopy \cite{jiang2024loopy} & 3.780 & 2.492 & 4.849 & 33.204 & 49.153 & 4.506 & 2.658 & 4.814 & 17.017 & 16.134 \\ 
    Hallo-3 \cite{hallo3} &
    3.451 & 2.257 & 3.933 & 38.481 & 42.125 & 4.006 & 2.462 & 4.448 & 28.840 & 26.029\\ 
    
     OmniHuman & \textbf{3.875} & \textbf{2.656} & \textbf{5.199} & \textbf{31.435} & \underline{46.393} & \textbf{4.564} & \textbf{2.815} & \textbf{5.255} & \textbf{16.970} & \textbf{15.906} \\ 
  \bottomrule
  \end{tabular}
        }
        {
            \caption{\textbf{Quantitative comparisons with audio-conditioned body animation baselines.}}
  \label{Tab2}
  \centering
  \begin{tabular}{cccccccc}
    \toprule
          Methods &IQA $\uparrow$ & ASE$\uparrow$ & Sync-C$\uparrow$ & FID$\downarrow$ & FVD$\downarrow$ &HKV& HKC$\uparrow$  \\
    \midrule
    DiffTED~\cite{diffted} & 2.701 & 1.703 & 0.926 & 95.455 & 58.871 & - & 0.769 \\
    DiffGest.~\cite{zhu2023taming}+MomicMo.~\cite{zhang2024mimicmotion} & 4.041 & 2.897 & 0.496 & 58.953 & 66.785  & 23.409  & 0.833  \\
    CyberHost~\cite{lin2025cyberhost} & 3.990 & 2.884 & 6.627  &  32.972 & 28.003 & 24.733 & 0.884 \\
    OmniHuman & \textbf{4.142} & \textbf{3.024} & \textbf{7.443} & \textbf{31.641} & \textbf{27.031} & 47.561 & \textbf{0.898} \\
  \bottomrule
  \end{tabular}
        }
     
\end{figure*}

\subsection{Ablation Studies on Omni-Conditions Training}

\textbf{Principles 1. } 
Here, we first analyze and explain Principle 1 of the omni-condition training. This principle guides us to introduce more training data using the weaker text condition compared to the commonly used audio and pose conditions in human animation tasks. In the upper part of Table~\ref{table:ablation}, we present the results of using varying proportions of text-conditioned data (T-DATA) on the results. It can be seen that more text-conditioned data improves core metrics such as FVD, FID, lipsync accuracy, and the richness and quality of gesture generation. We observed that when the amount of text-conditioned data is insufficient, the lipsync metric declines. This may be because a small amount of new data disrupts audio learning rather than enhancing text condition capability. As the data increases, the model leverages the shared backbone to learn better joint modality modeling, resulting in improved generated results. We also found that with continued training and increased data, the model gradually generates results that follow the quality of the input images rather than the distribution quality of the training set. Interestingly, while the image quality score may appear to decrease, the model actually learns to adhere to the distribution of the input images rather than the higher quality, limited audio-conditioned training data distribution. The introduction of text-conditioned data not only improves metrics but also significantly enhances the model's generalization to different input types. 

\begin{figure}[h]
    \centering
    \includegraphics[width=0.48\textwidth]{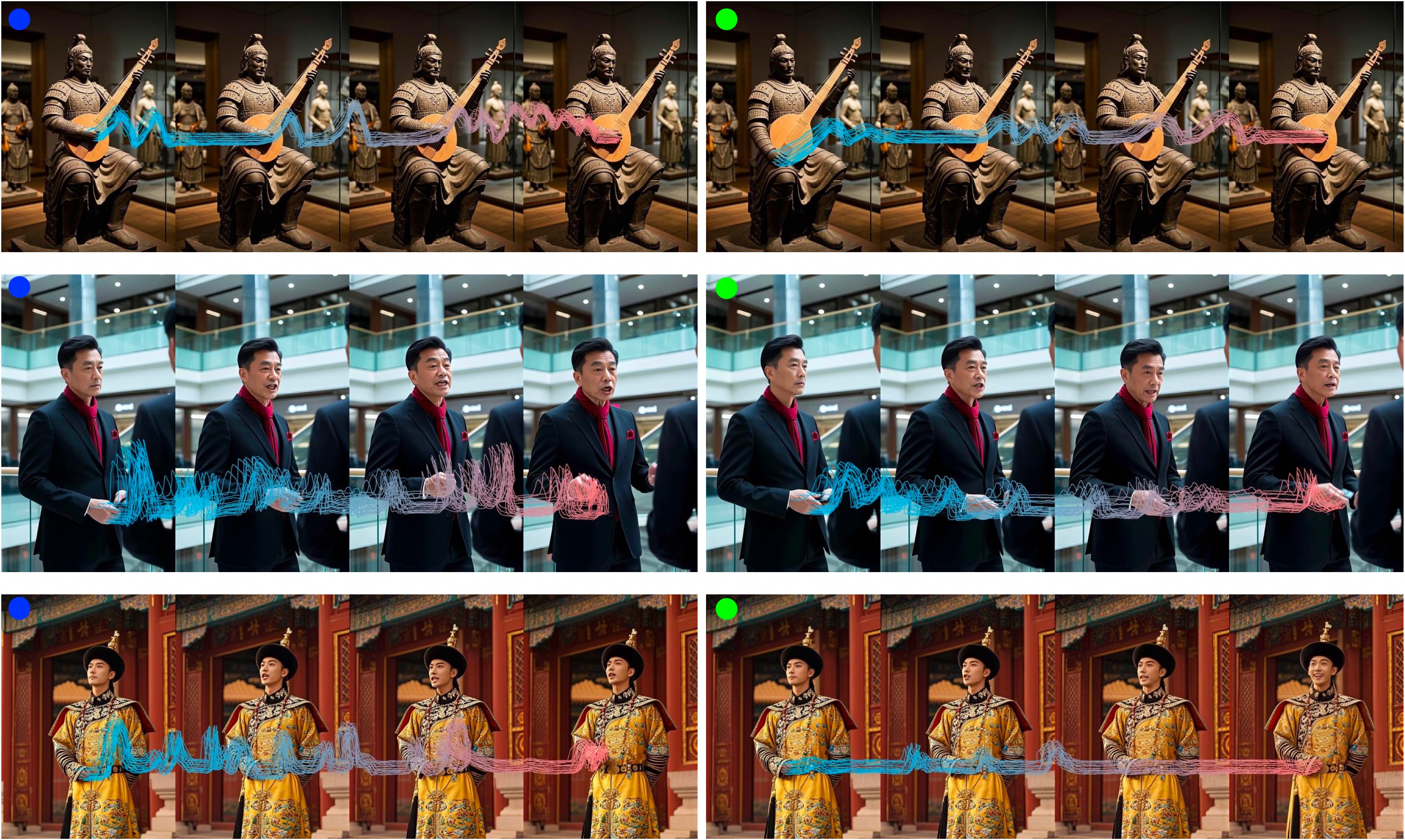}
    \caption{\small \textbf{Visual analysis of the hybrid-driven training model.} The gradient curves visualize the hand motion trajectories in the video. \textcolor{Color1}{Blue circle} denotes \textbf{IA} model results and \textcolor{Color2}{green circle} denotes \textbf{IAP} model results.}
    \label{fig:iapvsipa}

\end{figure}

\textbf{Principles 2.} This principle guides us on how to appropriately handle the relationships between different modality conditions during training, ensuring they complement each other to achieve satisfactory results. Since the sequential introduction of text and image in training is a common practice in the field of video generation, we follow this setup. However, the order of introducing pose and audio during training differs. Previous works~\cite{wang2024vexpress,jiang2024loopy,lin2025cyberhost,EchomimicV2} have introduced pose or pose-related features alongside audio features in the training process to stabilize it. By adhering to Principle 1, OmniHuman leverages a large amount of text-conditioned data to learn general video generation capabilities. We found that introducing pose too early in the training process can actually harm the model's ability to generate high-quality motion. 
For convenience, we use \textbf{I}, \textbf{T}, \textbf{A}, and \textbf{P} to represent the modalities of reference image, text, audio, and pose, respectively and their training order.
We use \textbf{IA} to denote our pure audio-driven model. 
In the lower part of Table~\ref{table:ablation}, we observe that training with pose first (\textbf{IPA}) leads to a comprehensive decline in the model's final output quality across multiple dimensions, affecting the upper limit of the model's performance.
Conversely, introducing pose later in the training process (\textbf{IAP}) does not harm the model's results, achieving comparable metrics and enabling the model to support both independent and mixed audio and pose driving. 
As shown in Figure~\ref{fig:iapvsipa}, we also found that the hybrid-driven training model \textbf{IAP} decouple hand movements from the associated audio. This mitigates the issue of excessively exaggerated hand movements in the generated videos, significantly enhancing the naturalness of human dynamics.

Based on Principle 2, we maintain a lower training proportion for pose compared to audio (\textbf{A\textgreater P, A=50\% and P=25\%}), by adjusting the condition drop rate. The results in Table~\ref{table:ablation} show that using the opposite training strategy (\textbf{A\textless P, A=50\% and P=75\%}) leads to degraded generation quality. We also validated the effects of different training proportions of audio compared to text in Supplementary Section 5. Specifically, we kept the text-condition ratio \textbf{T=90\%} constant and experimented with \textbf{A=10\%}, \textbf{A=50\%}, and \textbf{A=90\%}. According to the quantitative and qualitative results, the \textbf{A=50\%} ratio yields a more balanced performance, showing satisfactory generalization and lip-sync accuracy.

\begin{figure}[t]
    \centering
    \includegraphics[width=0.48\textwidth]{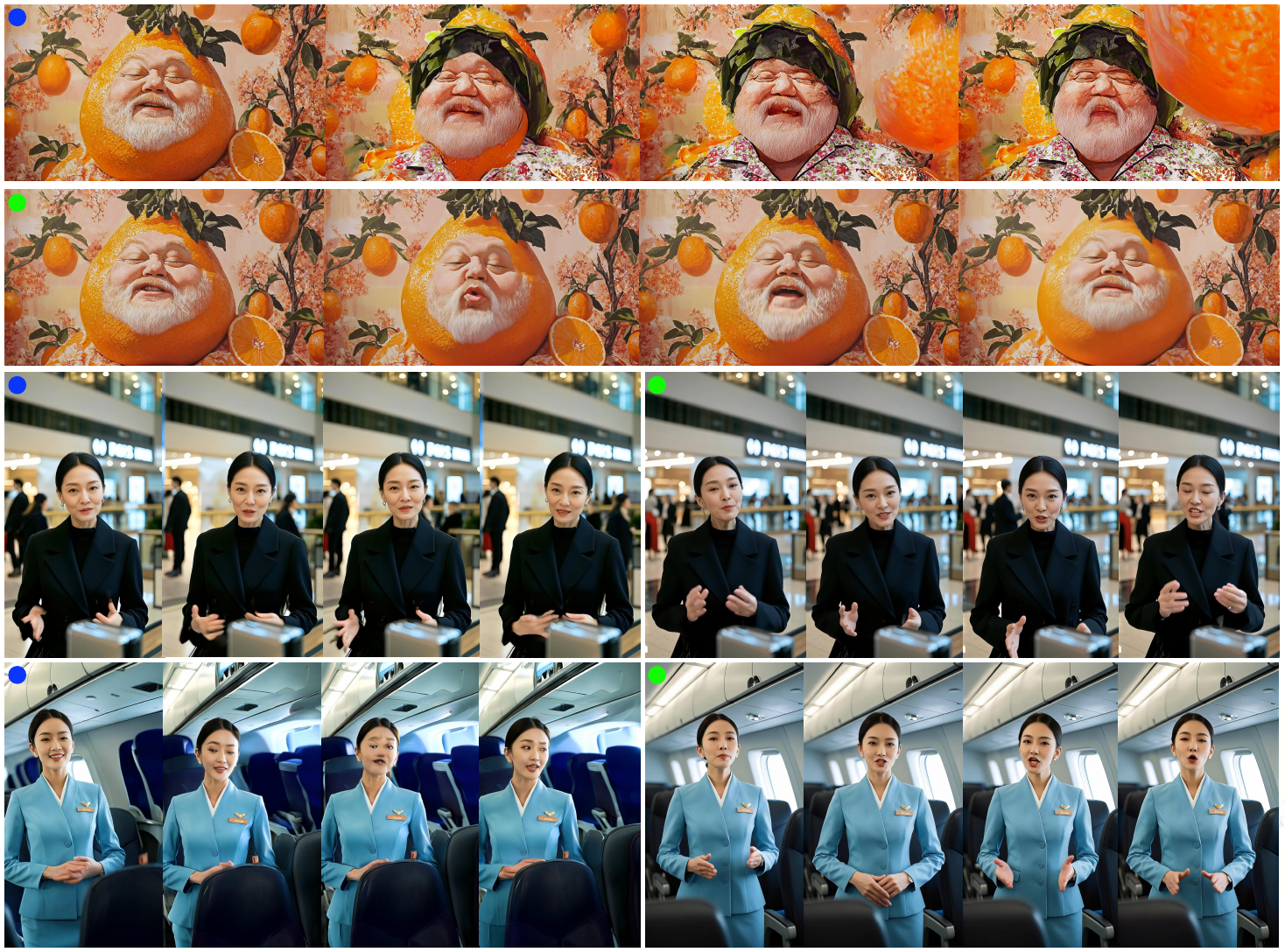}
    \caption{\small  \textbf{Visual comparison with the pretrained image-to-video model.} \textcolor{Color1}{Blue circle} denotes the I2V model and \textcolor{Color2}{green circle} denotes the OmniHuman.}
    \label{fig:t2v_vs_omni}
\end{figure}

\textbf{Effects of omni-condition training.} 
In addition to our investigation into training principles, we also provide a comparison with the results of  pretrained image-to-video model, as shown in Figure~\ref{fig:t2v_vs_omni}. The proposed omni-condition training allows our model to leverage the relationships between mixed conditions during inference. This significantly enhances the generative capabilities of pretrained I2V models, improving the quality of local human features such as gestures, as well as the overall realism and naturalness of movements in the generated videos. These results demonstrate that the proposed methods are crucial for improving generative performance.

\subsection{Comparisons with Existing Methods}
In this section, we provide a comparison between OmniHuman and existing methods, including audio-driven methods and pose-driven ones.

\textbf{Comparison with audio-driven methods.}
As shown in Tables \ref{Tab1} and \ref{Tab2}, OmniHuman demonstrates superior performance compared to leading specialized models in both portrait and body animation tasks using a single model. For audio-driven animation, the generated results cannot be identical to the original video, especially when the reference image contains only a head. The model's varying preferences for motion styles across different scenarios complicate performance measurement using a single metric. OmniHuman excels across almost all metrics in different datasets and tasks. Notably, existing methods target specific body proportions (portrait, half-body) with fixed input sizes and ratios. In contrast, OmniHuman supports various input sizes, ratios, and body proportions with a single model, achieving satisfactory and realistic results. More than just performance superiority, as far as we know, OmniHuman is the first solution capable of this, which further underscores the effectiveness of the proposed omni-condition training strategy. 

\textbf{Comparison with pose-driven methods.}
In addition to audio-driven animation, we also provide a comparison of OmniHuman with recent pose driving methods. It is worth noting that these comparison methods are all specifically trained for pose-driven tasks, while OmniHuman, being a unified model can handle different tasks including both audio-driven and pose-driven human video generation. The results in Table~\ref{table:V2V_SOTA} indicate that OmniHuman can also perform well on pose-only driving tasks with satisfactory visual quality and driving accuracy.

\begin{table}[t]
    \small
        \footnotesize
        \setlength{\tabcolsep}{5pt}
        \centering
        \caption{\small Comparison with video-driven body animation methods.}
    \begin{tabular}{l | c c c c c} 
        \toprule
        Methods  &IQA$\uparrow$ &ASE$\uparrow$  & FID$\downarrow$  & FVD$\downarrow$   & AKD$\downarrow$ \\ 
        \midrule
        DisCo~\citep{Disco} &3.707 &2.396 & 57.12 &64.52 & 9.313 \\
        AnimateAnyone~\citep{aa} &3.843 &2.718  & 26.87 &37.67 & 5.747\\
        MimicMotion~\citep{zhang2024mimicmotion}  &3.977 &2.842  &23.43 &22.97   & 8.536 \\
        CyberHost~\cite{lin2025cyberhost} &4.087 &2.967  &20.04 &7.7178  &3.123\\
        \midrule
        OmniHuman-1 &\textbf{4.111 } &\textbf{2.986} &\textbf{19.504} &\textbf{7.3184}  &\textbf{2.136}\\
        \bottomrule
    \end{tabular}
    \vspace{-0.05in}
  \label{table:V2V_SOTA}
\end{table}

\begin{figure}[tbp]
        \centering
    \includegraphics[width=0.48\textwidth]{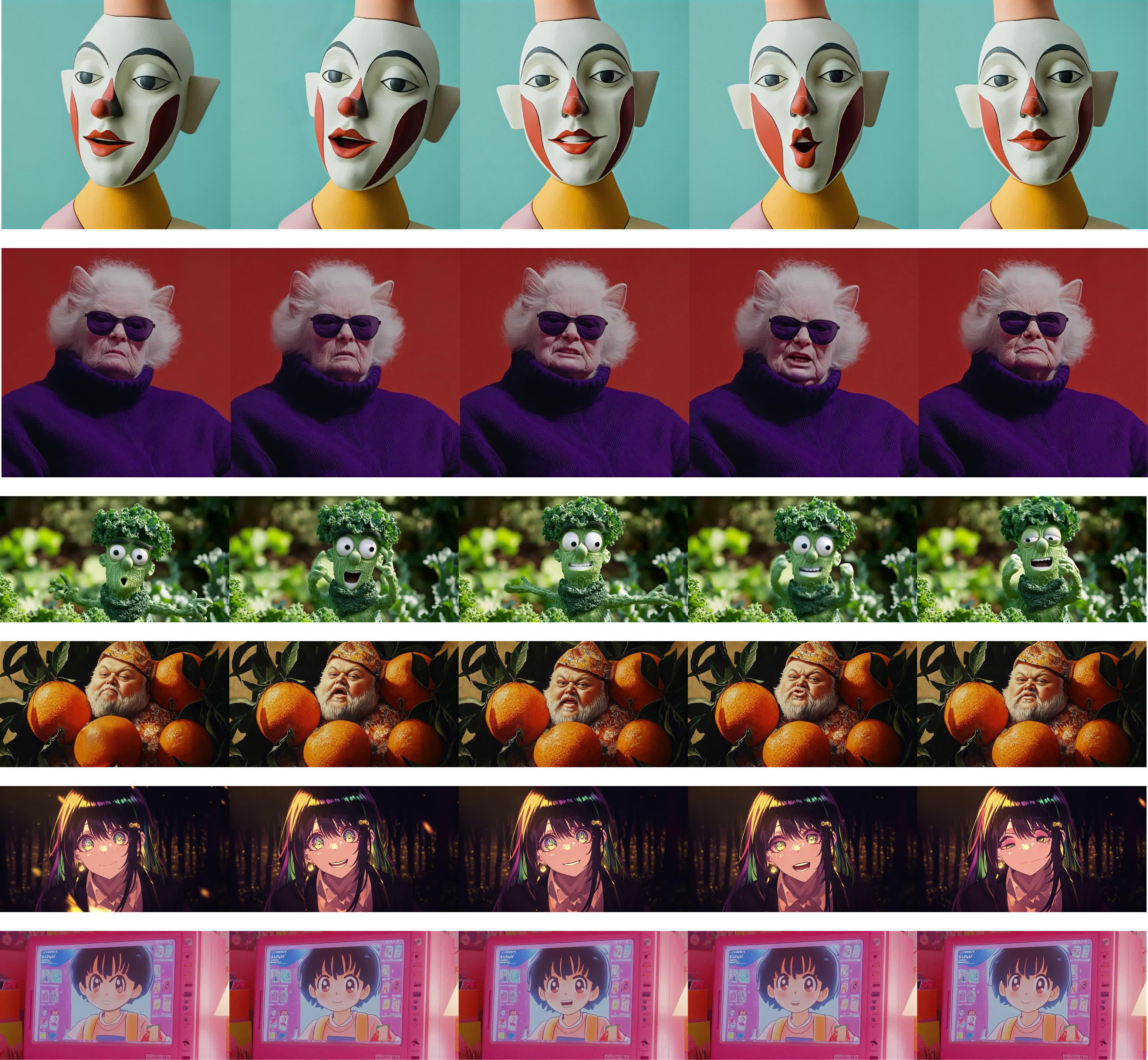}
    \caption{\small \textbf{The videos generated by OmniHuman based on input audio and images.} OmniHuman is compatible with stylized humanoid and 2D cartoon characters, and can even animate non-human images in an anthropomorphic manner.
}
    \label{fig:openset}
\end{figure}

\subsection{Extended Visual Results and Applications}
In Figure \ref{fig:openset}, we showcase OmniHuman's robust capabilities in human animation, which are challenging to quantify through metrics alone. OmniHuman is compatible with diverse input images and maintains the motion style of the input, such as preserving characteristic mouth movements in anime. Moreover, it excels in object interaction, generating videos of singing while playing instruments and natural gestures while holding objects. By adjusting the regions affected by CFG, OmniHuman can specify the speaker in the video, greatly expanding its application value. This versatility stems from OmniHuman's incorporation of various conditions and data during training, enabling it to learn general human animation through data scaling, unlike existing methods limited by narrowly filtered training data.

\section{Conclusion}
We propose OmniHuman, an end-to-end multimodality-conditioned human video generation framework that generates human videos based on a single image and motion signals (e.g., audio, pose, or both). OmniHuman employs a mixed data training strategy with multimodality motion conditioning, leveraging the scalability of mixed data to overcome the scarcity of high-quality data faced by previous methods. It outperforms existing approaches, producing highly realistic human videos from weak signals, especially audio. OmniHuman supports images of any aspect ratio (portraits, half-body, or full-body), delivering lifelike, high-quality results across various scenarios.

\section*{Limitation}
OmniHuman mitigates overfitting to audio variations through mixed-conditions training, especially by incorporating pose conditions. However, due to the weak correlation between audio and motion, uncoordinated or overly expressive movements still occur and object interactions can sometimes appear unrealistic. These stem from insufficient training data because when input images differ significantly from the training distribution, the model tends to generate unnatural results. To maintain synthesis stability, a relatively high CFG scale is needed, which can lead to some degree of overfitting. Future work will focus on adding richer motion conditions such as styles, intensities, and intentions to improve the naturalness of the generated motions.
\section*{Acknowledgments}
We thank Ceyuan Yang, Zhijie Lin, Yang Zhao, and Lu Jiang for their discussions and suggestions.

{
    \small
    \bibliographystyle{ieeenat_fullname}
    \bibliography{main}
}

\end{document}